\documentclass[runningheads]{llncs}
\usepackage{times}
\usepackage{hyperref}
\hypersetup{hidelinks,colorlinks=true,bookmarksnumbered=true,linkcolor=blue}
\usepackage{booktabs}
\usepackage[T1]{fontenc}
\def\doi#1{\href{https://doi.org/\detokenize{#1}}{\url{https://doi.org/\detokenize{#1}}}}
\usepackage{bm}
\usepackage{graphicx}
\usepackage{amssymb}
\usepackage{cite}
\usepackage{makecell}
\usepackage{times}
\usepackage{multirow}
\usepackage{makecell}
\usepackage{amsmath}
\usepackage[ruled,linesnumbered]{algorithm2e}

\usepackage{todonotes}

\usepackage{listings}
\begin{document}
\title{A Progressive Single-Modality to Multi-Modality Classification Framework for Alzheimer’s Disease Sub-type Diagnosis}

%
%
\newcommand*\samethanks[1][\value{footnote}]{\footnotemark[#1]}
\author{Yuxiao Liu\inst{1}\thanks{Co-first authors.}, Mianxin Liu\inst{2}\samethanks, Yuanwang Zhang\inst{1}, Kaicong Sun\inst{1}, Dinggang Shen\inst{1,3,4}}

\institute{School of Biomedical Engineering, ShanghaiTech University, Shanghai 201210, China\and
Shanghai Artificial Intelligence Laboratory, Shanghai 200232, China\and
Shanghai United Imaging Intelligence Co., Ltd., Shanghai 200232, China\and
Shanghai Clinical Research and Trial Center, Shanghai, 201210, China
}

\maketitle              
\vspace{-0.5cm}
\begin{abstract}

The current clinical diagnosis framework of Alzheimer’s disease (AD) involves multiple modalities acquired from multiple diagnosis stages, each with distinct usage and cost. Previous AD diagnosis research has predominantly focused on how to directly fuse multiple modalities for an end-to-end one-stage diagnosis, which practically requires a high cost in data acquisition. Moreover, a significant part of these methods diagnose AD without considering clinical guideline and cannot offer accurate sub-type diagnosis. In this paper, by exploring inter-correlation among multiple modalities, we propose a novel progressive AD sub-type diagnosis framework, aiming to give diagnosis results based on easier-to-access modalities in earlier low-cost stages, instead of modalities from all stages. Specifically, first, we design 1) a text disentanglement network for better processing tabular data collected in the initial stage, and 2) a modality fusion module for fusing multi-modality features separately. Second, we align features from modalities acquired in earlier low-cost stage(s) with later high-cost stage(s) to give accurate diagnosis without actual modality acquisition in later-stage(s) for saving cost. Furthermore, we follow the clinical guideline to align features at each stage for achieving sub-type diagnosis. Third, we leverage a progressive classifier that can progressively include additional acquired modalities (if needed) for diagnosis, to achieve the balance between diagnosis cost and diagnosis performance. We evaluate our proposed framework on large diverse public and in-home datasets (8280 in total) and achieve superior performance over state-of-the-art methods. Our codes will be released after the acceptance.

\end{abstract}

\section{Introduction}
\label{sec:introduction}
Alzheimer's disease (AD) poses a growing global health challenge, with an escalating number of subjects affected by cognitive impairment. The total healthcare costs for the treatment of AD are estimated as \$305 billion in 2023, which brings a heavy burden to society \cite{wong2020economic}. Accurate diagnosis allows effective treatment and hence alleviates AD-related burden. For AD diagnosis, key modalities encompass cognitive assessment and plasma biomarkers (recorded as tabular data), Magnetic Resonance Imaging (MRI), and Positron Emission Tomography (PET). The tabular data is relatively easier to access and thus can serve as initial screening data. For further diagnosis, MRI may be used to assess essential information regarding brain atrophy. When needed, PET (that can monitor the deposition of disease-related proteins in the brain) may be used to provide more specific evidence for AD diagnosis. Based on these modalities, deep learning (DL)-based methods usually can achieve higher diagnosis performance than the traditional ones. However, there still exist three major limitations in the existing DL-based methods. 

\textbf{First}, many DL-based studies build models for an end-to-end one-stage diagnosis using multiple modalities \cite{qiu2022multimodal, cheng2017cnns}. This can significantly increase the diagnosis cost because a considerable number of subjects can be firmly diagnosed in the early stages without additional modalities acquired in later stages.  This practical challenge limits the application of these studies in clinic.
\textbf{Second}, many DL-based works use only imaging data. Although there has emerged a series of works \cite{qiu2022multimodal,polsterl2021combining} using relatively easier-to-access tabular data for diagnosis, they simply input tabular data into the model without considering semantic information of these tabular values. Some works~\cite{liu2023clip,pellegrini2023xplainer} try to use pre-trained language encoder to encode textualized tabular data, but they need to use detailed templates to textualize tabular data, which may inadvertently introduce more common template information, instead of specific tabular data information.
\textbf{Finally}, previous methods directly classify the state of health and general AD, rather than align modality data with AD sub-types suggested by the clinical guideline \cite{cheng2017cnns, huang2019diagnosis, liu2014inter}. Consequently, these methods have difficulty in giving AD sub-type diagnosis results, and cannot assist in precise AD treatment in clinic.

Therefore, in this paper, we propose a progressive AD diagnosis framework (shown in Fig. \ref{wholeNetwork}) for cost-performance balance AD sub-type classification, which can effectively address the above-mentioned limitations. We summarize our contributions as follows:
\begin{itemize}
    \item  We propose a multi-modality and multi-stage contrastive framework for AD sub-type diagnosis, which can effectively align different modalities. This alignment ensures feature(s) acquired in earlier-stage(s) to also carry information from later stage(s), resulting in fewer diagnosis stages,  reducing diagnosis stages and costs.
    \item To better exploit clinical tabular data, we employ a text disentanglement network. This network extracts both common (from template) and specific features (from tabular data) from textualized tabular data. It minimizes the similarity of specific features to let them carry different aspects of subject information. Moreover, we use a modality fusion module to fuse and enhance feature representation across different stages (if needed) for better multi-modality diagnosis performance.
    \item We align features in each stage with AD sub-type criteria in the clinical guideline. Moreover, based on the alignments, we develop a confidence-driven multi-stage progressive diagnosis framework. The framework automatically determines whether to give a final diagnosis based on already collected data (without further acquisition to save cost) or to further acquire a more advanced modality in the next stage. 
    \item Experimental results on large diverse public and in-home datasets (8280 in total) show that our framework outperforms the other representative methods significantly.
\end{itemize}

\vspace{-0.2cm}
\section{Methods}
\vspace{-0.2cm}
\subsection{Notation \& Problem Formulation}
In alignment with real clinical procedure (often starting with easy-to-access data and later adding more difficult/expensive-to-access data if needed), our diagnostic process thus comprises $three\ distinct\ stages$, i.e., (Stage 1) using only tabular data, (Stage 2) using both tabular data and MRI data, and (Stage 3) using all the tabular data, MRI data, and PET data. Note that the early stage has a larger number of subjects for training, while the later stages always have less number of subjects for training (since most subjects have been firmly diagnosed in the early stage(s)). 

For each of the stages, a classifier for AD diagnosis is required. This classifier generates two options based on the diagnosis confidence, $either$ a confirmed decision $or$ a demand for next-stage modality acquisition. The options rely solely on how these acquired modalities are matched with the corresponding sub-type criteria in the guideline. In the final stage, the model is compelled to generate the diagnosis decision, because there will be no additional stages. Our comprehensive framework comprises the following parts. 
\textbf{Multi-modality Encoder}: This part elucidates the details on multi-modality feature extraction using the corresponding encoder across stages and multi-modality fusion for diagnosis (Detailed in Section \ref{encoder}). \textbf{Multi-modality Alignment and Clinical Guideline Alignment}: We show modality feature alignment among different stages and also the alignment between the fused feature with the corresponding sub-type criteria in the clinical guideline (Detailed in Section \ref{alignment}). \textbf{Progressive Classifier}: This part describes progressive diagnosis by referring to the diagnosis confidence (Detailed in Section \ref{sec:classifier}). 

\begin{figure*}[htbp]
  \centerline{
  \includegraphics[width=0.8\linewidth]{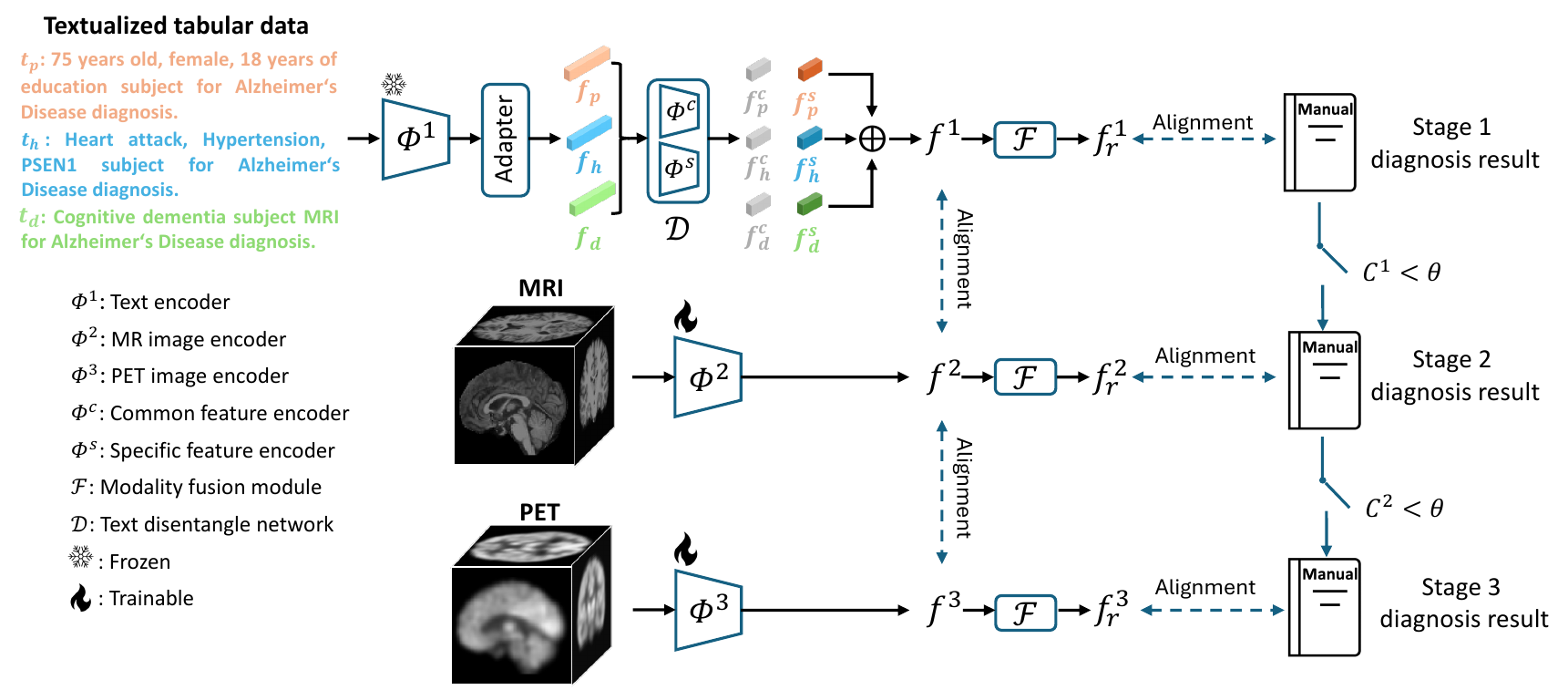}
  }
  \caption{
  We propose a progressive multi-modality and multi-stage contrastive framework for AD sub-type diagnosis. Starting from Stage 1, we acquire the features from the textualized tabular data, and align them with the corresponding sub-type criteria in the guideline. In cases where the diagnosis confidence is low, we progressively utilize MR and PET images in Stages 2 and 3, respectively. To acquire modality information of the later stage(s) without actual data acquisition, we further align the features from early stages with those from the later stages.}
  \label{wholeNetwork}
\end{figure*}

\subsection{Multi-modality Encoder}
\label{encoder}
We have three encoders $\Phi^1$, $\Phi^2$, and $\Phi^3$ in three corresponding stages for processing tabular data, MR images, and PET images, respectively.  At Stage $k$, we acquire the feature vector $f^k$ from the corresponding modality. Specifically, in Stage 1, we use the BioMedCLIP \cite{zhang2023large} text encoder as $\Phi^1$. For Stages 2 and 3, we train 3D image encoders $\Phi^2$ and $\Phi^3$ from scratch, because the BioMedCLIP utilizes the 2D image encoder which performs poorly for 3D brain image classification.

To better accommodate $\Phi^1$ to our task, we add an Adapter as shown in Fig. \ref{wholeNetwork}. We first textualize different tabular data into corresponding texts, including personal text $t_p$, healthy text $t_h$, and dementia text $t_d$, following the same text template (with more details are given in Sec. \ref{template_design}). Then, we separately input them into $\Phi^1$. 
Each output from the $\Phi^1$ is forwarded into the Adapter, which is a 4-layer multi-layer perceptron (MLP), and mapped to the corresponding text feature vector $f_*\in$\{$f_p$, $f_h$, $f_d$\}. 

Previous research~\cite{liu2023clip} has demonstrated that a detailed text template can make the language encoder perform better on downstream tasks. However, a highly detailed template may inadvertently introduce more generic information from the template itself rather than specific information from diverse tabular data. This can make the language encoder only acquire similar features for different textualized tabular data and thus hard to distinguish these features \cite{seibold2022breaking}. To address this issue, we construct a text disentanglement network $\mathcal{D}$ as formulated below.
\vspace{-0.5cm}
\subsubsection{Text disentanglement network.}
\label{sec:text_disentangle}
The text feature vectors ($f_p$, $f_h$, $f_d$) output by $\Phi_1$ are jointly forwarded into the disentanglement network $\mathcal{D}$, as shown in Fig. \ref{wholeNetwork}. It aims to disentangle each feature vector $f_*\in\{f_p, f_h, f_d\}$ into common a feature vector $f^c_*$ (from template) and specific feature vector $f^s_*$ (from tabular data) \cite{zhang2021disentangled,balasubramanian2020polarized}. 
Specifically, we employ a common encoder $\Phi^c$ and a specific encoder $\Phi^s$ to map $f_*$ into $f_*^c$ and $f_*^s$, respectively. We minimize their mutual information (MI) with loss $\mathcal{L}_\mathcal{D}^{MI}=\sum_{f_* \in\left\{f_p, f_h, f_d\right\}} M I\left(f^c_*, f^s_*\right)$. 
Besides, it is crucial to ensure features to exhibit similarity across different text features. Conversely, the specific features carry different aspects of subject information and should be orthogonal with each other. 
Therefore, we maximize the similarity of common features and minimize the similarity of specific features in all paired text features. Herein, we adopt the orthogonal loss $\mathcal{L}_\mathcal{D}^O$ as formulated below,
\begin{equation}
    \mathcal{L}_\mathcal{D}^O=
    \frac{1}{2} \times \sum_{\substack{f_* \in \\
\left\{f_p, f_h, f_d\right\}}} 
\sum_{\substack{f_{**} \in \\
\{f_p, f_h, f_d\}}}-
\cos \left(f_*^c, f_{**}^c\right)+\left|\cos \left(f_*^s, f_{**}^s\right)
\right|.
\end{equation}
Finally, we concatenate disentangled features of each text feature, namely $[f_*^c,f_*^s]$, $f_*\in \{f_p,f_h,f_d\}$ as each output of $\mathcal{D}$ to maximize the utilization of information from both the template and tabular data. Subsequently, all the output features from $\mathcal{D}$ are further concatenated to obtain features of the textualized tabular data, feature $f^1$, at Stage 1.

\subsubsection{Multi-modality fusion.}
To adaptively fuse multi-modality features from multiple stages ($f^1$, $f^2$, and $f^3$) for final diagnosis, we apply a transformer-based multi-modality fusion module $\mathcal{F}$ as denoted in Fig. \ref{wholeNetwork}. The fusion module $\mathcal{F}$ takes a transformer architecture to adaptively fuse $f^1$, $f^2$, and $f^3$ into a fixed-length vector for the diagnosis. Specifically, $\mathcal{F}$ first concatenates $k$ available feature(s) at Stage $k$ into a vector $f^{1:k}=[f^1,...f^k], k\leq3$. Then, $f^{1:k}$ is further concatenated with learnable query $Q$ as key and value, namely $K=V=[f^{1:k},Q]$. Finally, the vectors $Q$, $K$, and $V$ are used to calculate the attention $ATT$ and feed forward block $FFW$ \cite{vaswani2017attention} (an MLP with a ReLU) to finally acquire $f_r^k$ as below:
\begin{equation}
    f_r^k = Q+FFW\left(Q+ATT(Q,K,V)\right).
    \label{eq:feature}
\end{equation}
Based on the $ATT$ and the $FFW$, $f_r^k$ is mapped to the same dimension as $Q$. Meanwhile, since our fusion module can process various lengths of vectors, we can train our $\mathcal{F}$ using all the subjects we have.
\subsection{Multi-modality Alignment and Clinical Guideline Alignment}
\label{alignment}
\subsubsection{Multi-modality alignment.}
Since multiple modalities are acquired from the same subject, there often exists underlying relevance among these modalities \cite{pan2018synthesizing}. To enhance diagnosis performance in earlier stages, we propose to let the early stage encoder(s) entail modality information of the later stage(s) without actual acquisition. Therefore, we attempt to align the modality features of all the stages. The alignment loss $\mathcal{L}_{alig}$ is formulated below:
\begin{equation}
    \mathcal{L}_{alig}=\sum_{k=1}^2 ||f^k-f^{k+1}||_2.
\end{equation}
Based on the alignment of features in previous stages with the later stage ones, the features of previous stages can embody particular information that only the later stages can provide. This can improve the diagnosis performance in the previous stages, and hence save diagnosis cost.
\vspace{-0.2cm}
\subsubsection{Clinical guideline alignment.}
To achieve sub-type diagnosis with $f_r^k$ at Stage $k$, we align $f_r^k$ with the sub-type criteria in the clinical guideline. 
To be specific, we encode both the corresponding and non-corresponding sub-type criteria of the subject, to make $f_r^k$ not only like its corresponding criteria $Y$ but also different from the non-corresponding criteria $-^Y$. We define the contrastive loss $\mathcal{L}_{con}^k$ as the negative logarithm of the corresponding criteria contrastive score as expressed below:
\begin{equation}
    \mathcal{L}_{con}^k = -\log\frac{\exp((f_r^k)\Phi^1(Y))}{\sum_{y\in ^-Y}\exp((f_r^k)\Phi^1(y)))+\exp((f_r^k)\Phi^1(Y))}.
    \label{s1_contrastive}
\end{equation}
During inference, we compare the contrastive score of all the AD sub-types and choose the largest one as the diagnosis decision.

\subsection{Progressive Classifier}
\label{sec:classifier}
In clinical practice, it is important to consider the cost of accurate diagnosis. If the previous stage(s) can make confident diagnosis decisions already, there will be no need for later stage(s). Therefore, our goal is to minimize the total costs to the maximum extent without compromising classification performance. To this end, inspired by \cite{zheng2023selected}, we define diagnosis confidence $C^k$ in each Stage $k$ by subtracting the second highest contrastive score from the highest one in clinical guideline alignment. This subtracted value indicates how confident is the network about the diagnosis decision. Then, we set a threshold $\theta$ on $C^k$. As we previously mentioned in Section \ref{sec:introduction}, there are two options for the classifier in Stage $k$. If diagnosis confidence $C^k$ is larger than the threshold $\theta$, it leads to the first option, i.e., to give the diagnosis result and stop. Otherwise, a further stage is required. Inspired by~\cite{trapeznikov2012multi}, we encourage the model to give confident diagnosis results in earlier stages. For each stage, we set the loss $\mathcal{L}^k$ to avoid postponement by multiplying the negative of the confidence by a penalty $\delta^k>0$.
\begin{equation}
    \mathcal{L}^k = \left\{ 
    \begin{aligned}
    &\mathcal{L}_{con}^k  &C^k\geq\theta^k \text{ or } k=3 \cr 
    &-C^k\delta^k  &C^k<\theta^k 
    \end{aligned}
\right.
\end{equation}

\section{Experiments}
\vspace{-0.2cm}
\subsection{Dataset}
\label{Dataset}
\vspace{-0.2cm}
We use data from four datasets including Alzheimer’s Disease Neuroimaging Initiative (ADNI) \cite{ADNIdataset1, ADNIdataset2}, Open Access Series of Imaging Studies (OASIS) \cite{lamontagne2019oasis}, National Alzheimer's Coordinating Center (NACC) \cite{beekly2007national}, and an in-house dataset from Huashan Hospital. We follow the IWG-2 diagnosis guideline \cite{dubois2014advancing} and group the subjects into four sub-types, including typical, atypical, pre-clinical AD, and normal control. The corresponding sub-type criteria are detailed in the supplementary material. We have totally 8280 subjects with tabular data, 4842 subjects with both tabular and MRI data, and 2336 subjects with all the tabular, MRI, and PET data. We trained all models on the NACC and ADNI datasets because these two datasets have the most complete data. Datasets are randomly divided into 5 equal folds. 3 folds are used for training and remaining 2 folds for validation and testing. Model is tested on subjects with full modalities. We do not divide the dataset with overlapping subjects to avoid the leakage. Note that although we use demetia information, this will not cause the data leakage as IWG-2 \cite{dubois2014advancing} considers both dementia information and in-vivo evidence for AD diagnosis instead of dementia information only.
\vspace{-0.3cm}
\subsection{Implementation Details}
To ensure a fair comparison, we use 3D CNN as image encoders (as detailed in supplementary material) for all comparison methods. We use the SGD optimizer with a learning rate of $10^{-4}$. We train our model for 100 epochs. We choose the MSE as the $\mathcal{L}_{alig}$. We set $\theta$ as 0.3, and weights for all the loss functions are 1. $\delta^1$ and $\delta^2$ are set to be 1 and 1.5, respectively. For text input $t_p$, we consider age, education, and gender. For $t_h$, we consider healthy information, including heart attack, hypertension, stroke, alcohol abuse, psychiatric disorder, and the blood test. For $t_d$, we employ dementia level. As some subjects may lack specific tabular information, we only input available textualized tabular data to the text encoder.
\vspace{-0.2cm}
\subsection{Results and Discussions}
We employ six distinct evaluation metrics to assess both diagnostic performance and cost-effectiveness for subjects with full modalities (totally 837 subjects). The first four metrics are for sub-type diagnosis and the other two are for cost-effectiveness. For estimating costs, we use the mean stage number as a proxy. 
Finally, we compute the ratio of AUC to the estimated cost to measure the cost-effectiveness. 

\begin{table}[]
    \centering
        \caption{Quantitative comparison with other methods on our test dataset.}
        \begin{tabular}{c|c|c|c|c|c|c}
        \hline Models & Acc & Spe & Sens & AUC & Cost& AUC/Cost Ratio\\
        \hline CoCoOp & 74.1 & 83.3 & 74.6 & 72.9 &2.80&26.03\\
        \hline Xplainer & 75.3 & 85.6 & 73.2 & 75.5&2.53&29.84 \\
        \hline Ours $w/o$ progressive classifier &80.8&87.1&80.6&81.7 &3&27.23 \\
        \hline Ours $w/o$ multi-modality alignment  &76.5&86.8&76.6&75.3 &2.45&30.73 \\
        \hline Ours $w/o$ $\mathcal{D}$ &74.5 &83.6 &71.6 &73.8 &2.42& 30.50\\
        \hline Ours $w/o$ $\mathcal{F}$ &75.8&84.9&76.2&73.7 &2.52&29.24 \\
        \hline Ours & 78.2 & 88.3 & 77.8 & 77.4&2.21&\textbf{35.02}\\
        \hline
        \end{tabular}
    \label{tab:comparison}
\end{table}

From Tab. \ref{tab:comparison}, we can observe that our framework demonstrates a superior classification performance over 
the representative methods (CoCoOp \cite{zhou2022conditional} and Xplainer \cite{pellegrini2023xplainer}), which are also built on multi-stage and multi-modality fusion. These two comparison methods use different text encoders, but the same image encoder as ours. Our performance improvement mainly arises from two reasons. $First$, we use text disentanglement network $\mathcal{D}$ to extract more information from different textualized tabular data. $Second$, we further propose multi-modality alignment to let early-stage features entail later-stage information, allowing the model to have improved diagnosis performance in the early stages for reducing acquisition cost. The ablation studies on $w/o$ $\mathcal{D}$ and multi-modality alignment further validate our conclusion. Besides, we also show effectiveness of the progressive classifier. We can see that, although there exists a certain performance drop in diagnosis accuracy compared to the one without using the progressive classifier, it is expected, since in earlier stages, we use fewer modalities for diagnosis. However, the use of the progressive classifier achieves a substantial reduction in diagnosis cost, and also improves the AUC/Cost ratio significantly, implying that our model has great potential for practical applications in clinical routine.
\vspace{-0.2cm}
\subsubsection{Text templates design.}
\label{template_design}
In our evaluation, we test three types of text templates as outlined below (by using age information in $t_p$ as an example). The first template (template 1) solely focuses on basic tabular information of the subject. For templates 2 and 3, we introduce more detailed descriptions. The corresponding performance is depicted in Table \ref{tab:prompts}. We can observe that, employing a more detailed text template can enhance performance. 
Hence, we choose to utilize template 3 in our method. Besides, we can see that the more detailed the template is, the more contribution $\mathcal{D}$ can make to diagnosis performance. This demonstrates that $\mathcal{D}$ can more effectively extract features from the detailed text templates. 
\begin{itemize}
    \item \textbf{Template 1:} \{75 years old\} 
    \item \textbf{Template 2:} \{75 years old\}  subject
    \item \textbf{Template 3:} \{75 years old\}  subject for Alzheimer's Disease diagnosis
\end{itemize}
\vspace{-0.3cm}
\begin{table}[]
    \centering
  
    \caption{Evaluation on different text templates and text disentanglement network $\mathcal{D}$.}
    \resizebox{0.98\textwidth}{!}{
        \begin{tabular}{c|c|c|c|c|c|c|c|c|c|c|c|c}
        \hline &\multicolumn{6}{c}{w/ $\mathcal{D}$} &\multicolumn{6}{|c}{w/o $\mathcal{D}$}\\ 
        \hline Template & Acc & Spe & Sens & AUC &Cost& AUC/Cost Ratio&Acc &Spe  &Sens &AUC &Cost & AUC/Cost Ratio\\
        \hline Template 1 &70.7 &78.3 &71.9 &71.5&2.39&29.91              &69.8&77.7 &70.5 &71.2 &2.46 &28.94\\
        \hline Template 2 &73.1 &80.9 &73.8 &73.5&2.35&31.27              &71.2&78.7 &70.9 &71.5 &2.48 &28.83\\
        \hline Template 3 &78.2 &88.3 &77.8 &77.4&2.21&35.02              &74.5&83.6 &71.6 &73.8 &2.42 &30.50\\
        \hline
        \end{tabular}
    }
    \label{tab:prompts}
\end{table}

\vspace{-0.3cm}
\subsubsection{Effect of threshold on confidence.}
We delve deeper into exploring the impact of using varying confidence thresholds in the progressive classifier. A relatively lower confidence threshold encourages the model to make decisions in earlier stage(s) and substantially decreases acquisition costs. However, it may also lead to incorrect diagnosis results that cannot be corrected by subsequent stage(s). On the contrary, opting for a higher confidence threshold can lead to more accurate diagnosis results but cause increased diagnosis costs. In Table \ref{tab:confidence_influence}, we present performance and cost ratio using different confidence thresholds. We set the same threshold value in stages 1 and 2. 
As we can see, as the threshold increases, diagnosis performance is also improved, but the improvement ratio slows down. With regard to the AUC/Cost ratio, it increases first and then decreases dramatically. This effect is reasonable since, when the threshold becomes too large, most of the subjects are required to finish later stages to obtain diagnosis decision. This will benefit diagnosis correctness, but with sacrifice of the AUC/cost ratio since many subjects can be firmly diagnosed in early stages without going to next expensive stage(s).
\vspace{-0.4cm}
\begin{table}[]
    \centering
        \caption{Evaluation on different confidence thresholds.}
        \begin{tabular}{c|c|c|c|c|c|c}

        \hline Threshold   & Acc & Spe &Sens & AUC &Cost&AUC/Cost ratio\\
        \hline 0.1       &71.8 &70.9 &72.3 &72.2 &2.09 &34.54\\
        \hline 0.3       &78.2 &88.3 &77.8 &77.4 &2.21 &35.02\\
        \hline 0.5        &78.9 &87.7 &79.3 &79.6 &2.63 &20.26\\
        \hline
        \end{tabular}
    \label{tab:confidence_influence}
\end{table}

\vspace{-0.8cm}
\section{Conclusion}
In this work, we present a multi-modal and multi-stage framework for cost-effective AD sub-type diagnosis. Different from the existing works, our framework is to progressively give diagnosis results from a single modality to multiple modalities. Specifically, in the first stage, we utilize only tabular data. To effectively extract textualized tabular information, we propose a text disentanglement network to disentangle common and specific features of different tabular data. If diagnosis confidence of the first stage is lower than a pre-defined threshold, the respective subject requires later stages, which integrates the MRI and PET data. To effectively fuse multiple modalities, we propose a transformer-based multi-modality fusion module. To improve diagnosis performance of early stages and reduce overall diagnosis costs, we propose to align features of different stages to allow early stages to contain particular information only available in later stage(s). Furthermore, we match modality features and the AD diagnosis guideline to achieve improved sub-type diagnosis. Extensive experiments demonstrate the superiority of our framework, showing promise in real AD diagnosis.
\bibliographystyle{plain}

\bibliography{ref} 

\begin{thebibliography}{10}

\bibitem{ADNIdataset2}
Paul~S. Aisen, Ronald~C. Petersen, Michael Donohue, and Michael~W. Weiner.
\newblock {Alzheimer's Disease Neuroimaging Initiative 2 Clinical Core: Progress and Plans}, 2015.

\bibitem{balasubramanian2020polarized}
Vikash Balasubramanian, Ivan Kobyzev, Hareesh Bahuleyan, Ilya Shapiro, and Olga Vechtomova.
\newblock Polarized-vae: Proximity based disentangled representation learning for text generation.
\newblock {\em arXiv preprint arXiv:2004.10809}, 2020.

\bibitem{beekly2007national}
Duane~L Beekly, Erin~M Ramos, William~W Lee, Woodrow~D Deitrich, Mary~E Jacka, Joylee Wu, Janene~L Hubbard, Thomas~D Koepsell, John~C Morris, Walter~A Kukull, et~al.
\newblock The national alzheimer's coordinating center (nacc) database: the uniform data set.
\newblock {\em Alzheimer Disease \& Associated Disorders}, 21(3):249--258, 2007.

\bibitem{cheng2017cnns}
Danni Cheng and Manhua Liu.
\newblock Cnns based multi-modality classification for ad diagnosis.
\newblock In {\em 2017 10th international congress on image and signal processing, biomedical engineering and informatics (CISP-BMEI)}, pages 1--5. IEEE, 2017.

\bibitem{dubois2014advancing}
Bruno Dubois, Howard~H Feldman, Claudia Jacova, Harald Hampel, Jos{\'e}~Luis Molinuevo, Kaj Blennow, Steven~T DeKosky, Serge Gauthier, Dennis Selkoe, Randall Bateman, et~al.
\newblock Advancing research diagnostic criteria for alzheimer's disease: the iwg-2 criteria.
\newblock {\em The Lancet Neurology}, 13(6):614--629, 2014.

\bibitem{huang2019diagnosis}
Yechong Huang, Jiahang Xu, Yuncheng Zhou, Tong Tong, Xiahai Zhuang, and Alzheimer’s Disease Neuroimaging~Initiative (ADNI).
\newblock Diagnosis of alzheimer’s disease via multi-modality 3d convolutional neural network.
\newblock {\em Frontiers in neuroscience}, 13:509, 2019.

\bibitem{ADNIdataset1}
Clifford~R. Jack.
\newblock {Magnetic Resonance Imaging in Alzheimer's Disease Neuroimaging Initiative 2}, 2015.

\bibitem{lamontagne2019oasis}
Pamela~J LaMontagne, Tammie~LS Benzinger, John~C Morris, Sarah Keefe, Russ Hornbeck, Chengjie Xiong, Elizabeth Grant, Jason Hassenstab, Krista Moulder, Andrei~G Vlassenko, et~al.
\newblock Oasis-3: longitudinal neuroimaging, clinical, and cognitive dataset for normal aging and alzheimer disease.
\newblock {\em MedRxiv}, pages 2019--12, 2019.

\bibitem{liu2014inter}
Feng Liu, Chong-Yaw Wee, Huafu Chen, and Dinggang Shen.
\newblock Inter-modality relationship constrained multi-modality multi-task feature selection for alzheimer's disease and mild cognitive impairment identification.
\newblock {\em NeuroImage}, 84:466--475, 2014.

\bibitem{liu2023clip}
Jie Liu, Yixiao Zhang, Jie-Neng Chen, Junfei Xiao, Yongyi Lu, Bennett A~Landman, Yixuan Yuan, Alan Yuille, Yucheng Tang, and Zongwei Zhou.
\newblock Clip-driven universal model for organ segmentation and tumor detection.
\newblock In {\em Proceedings of the IEEE/CVF International Conference on Computer Vision}, pages 21152--21164, 2023.

\bibitem{pan2018synthesizing}
Yongsheng Pan, Mingxia Liu, Chunfeng Lian, Tao Zhou, Yong Xia, and Dinggang Shen.
\newblock Synthesizing missing pet from mri with cycle-consistent generative adversarial networks for alzheimer’s disease diagnosis.
\newblock In {\em Medical Image Computing and Computer Assisted Intervention--MICCAI 2018: 21st International Conference, Granada, Spain, September 16-20, 2018, Proceedings, Part III 11}, pages 455--463. Springer, 2018.

\bibitem{pellegrini2023xplainer}
Chantal Pellegrini, Matthias Keicher, Ege {\"O}zsoy, Petra Jiraskova, Rickmer Braren, and Nassir Navab.
\newblock Xplainer: From x-ray observations to explainable zero-shot diagnosis.
\newblock {\em arXiv preprint arXiv:2303.13391}, 2023.

\bibitem{polsterl2021combining}
Sebastian P{\"o}lsterl, Tom~Nuno Wolf, and Christian Wachinger.
\newblock Combining 3d image and tabular data via the dynamic affine feature map transform.
\newblock In {\em Medical Image Computing and Computer Assisted Intervention--MICCAI 2021: 24th International Conference, Strasbourg, France, September 27--October 1, 2021, Proceedings, Part V 24}, pages 688--698. Springer, 2021.

\bibitem{qiu2022multimodal}
Shangran Qiu, Matthew~I Miller, Prajakta~S Joshi, Joyce~C Lee, Chonghua Xue, Yunruo Ni, Yuwei Wang, Ileana De~Anda-Duran, Phillip~H Hwang, Justin~A Cramer, et~al.
\newblock Multimodal deep learning for alzheimer’s disease dementia assessment.
\newblock {\em Nature communications}, 13(1):3404, 2022.

\bibitem{seibold2022breaking}
Constantin Seibold, Simon Rei{\ss}, M~Saquib Sarfraz, Rainer Stiefelhagen, and Jens Kleesiek.
\newblock Breaking with fixed set pathology recognition through report-guided contrastive training.
\newblock In {\em International Conference on Medical Image Computing and Computer-Assisted Intervention}, pages 690--700. Springer, 2022.

\bibitem{trapeznikov2012multi}
Kirill Trapeznikov, Venkatesh Saligrama, and David Casta{\~n}{\'o}n.
\newblock Multi-stage classifier design.
\newblock In {\em Asian conference on machine learning}, pages 459--474. PMLR, 2012.

\bibitem{vaswani2017attention}
Ashish Vaswani, Noam Shazeer, Niki Parmar, Jakob Uszkoreit, Llion Jones, Aidan~N Gomez, {\L}ukasz Kaiser, and Illia Polosukhin.
\newblock Attention is all you need.
\newblock {\em Advances in neural information processing systems}, 30, 2017.

\bibitem{wong2020economic}
Winston Wong.
\newblock Economic burden of alzheimer disease and managed care considerations.
\newblock {\em The American journal of managed care}, 26(8 Suppl):S177--S183, 2020.

\bibitem{zhang2021disentangled}
Jingfeng Zhang, Haiwen Hong, Yin Zhang, Yao Wan, Ye~Liu, and Yulei Sui.
\newblock Disentangled code representation learning for multiple programming languages.
\newblock {\em Findings of the Association for Computational Linguistics: ACL-IJCNLP 2021}, 2021.

\bibitem{zhang2023large}
Sheng Zhang, Yanbo Xu, Naoto Usuyama, Jaspreet Bagga, Robert Tinn, Sam Preston, Rajesh Rao, Mu~Wei, Naveen Valluri, Cliff Wong, et~al.
\newblock Large-scale domain-specific pretraining for biomedical vision-language processing.
\newblock {\em arXiv preprint arXiv:2303.00915}, 2023.

\bibitem{zheng2023selected}
Zefeng Zheng, Shaohua Teng, Naiqi Wu, Luyao Teng, Wei Zhang, and Lunke Fei.
\newblock Selected confidence sample labeling for domain adaptation.
\newblock {\em Neurocomputing}, 555:126624, 2023.

\bibitem{zhou2022conditional}
Kaiyang Zhou, Jingkang Yang, Chen~Change Loy, and Ziwei Liu.
\newblock Conditional prompt learning for vision-language models.
\newblock In {\em Proceedings of the IEEE/CVF Conference on Computer Vision and Pattern Recognition}, pages 16816--16825, 2022.

\end{thebibliography}

\end{document}


\title{Supplementary Material for A Progressive Single-Modality to Multi-Modality Classification Framework for Alzheimer’s Disease Sub-type Diagnosis}
\author{ Paper ID: 1636
}

\maketitle     
\section{CNN encoder}
\begin{figure}[htbp]
      \centerline{
  \includegraphics[width=1\linewidth]{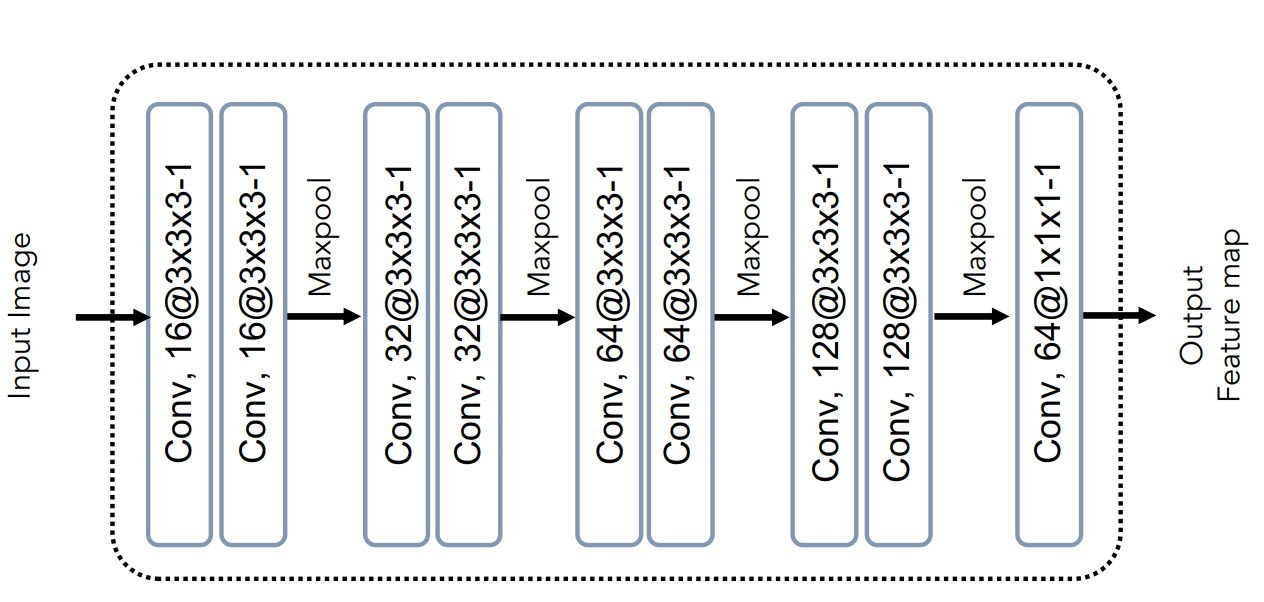}
  }
    \caption{CNN encoder for processing 3D brain images.}
    \label{CNN_encoder}
\end{figure}
\section{AD sub-type criteria input to the language encoder}
\begin{itemize}
    \item Criteria for typical AD:  Significant episodic memory impairment and In-vivo evidence of Alzheimer’s disease;
    \item Criteria for atypical AD: Posterior or logopenic or frontal of Alzheimer’s disease and In-vivo evidence of Alzheimer’s disease;
    \item Criteria for  pre-clinical AD: Absence of specific clinical phenotype and In-vivo evidence of Alzheimer’s disease 
    \item Normal control: Absence of specific clinical phenotype and No evidence of Alzheimer’s disease 
    
\end{itemize}